%% file: main.tex
\documentclass[twocolumn]{article}

\usepackage{framed,multirow}

\usepackage{amssymb}
\usepackage{amsmath}
\usepackage{latexsym}
\usepackage{natbib}
\usepackage{authblk}
\usepackage[hyphens,spaces,obeyspaces]{url}
\usepackage{xcolor}
\definecolor{newcolor}{rgb}{.8,.349,.1}

\usepackage[utf8]{inputenc}
\usepackage{hyperref}
\usepackage{multicol}
\usepackage{adjustbox}
\usepackage{lipsum}

\usepackage{geometry}

\let\svthefootnote\thefootnote
\newcommand\freefootnote[1]{%
  \let\thefootnote\relax%
  \footnotetext{#1}%
  \let\thefootnote\svthefootnote%
}

\title{A survey on bias in visual datasets}
\date{\today}

\author[1,3]{Simone Fabbrizzi\thanks{Corresponding author: simone.fabbrizzi@iti.gr}}
\author[1]{Symeon Papadopoulos}
\author[2]{Eirini Ntoutsi}
\author[1]{Ioannis Kompatsiaris}
\affil[1]{CERTH-ITI, Thessaloniki, Greece}
\affil[2]{Freie Universit\"at, Berlin, Germany }
\affil[3]{Leibniz Universit\"at, Hannover, Germany }

\begin{document}
\maketitle

\freefootnote{This work is currently under review at Computer Vision and Image Understanding.}

\begin{abstract}
Computer Vision (CV) has achieved remarkable results, outperforming humans in several tasks. Nonetheless, it may result in significant discrimination if not handled properly as CV systems highly depend on the data they are fed with and can learn and amplify biases within such data. 
Thus, the problems of understanding and discovering biases are of utmost importance. Yet, there is no comprehensive survey on bias in visual datasets.
Hence, this work aims to: i) describe the biases that might manifest
in visual datasets; ii) review the literature on methods for bias discovery and quantification in visual datasets; iii) discuss existing attempts to collect bias-aware visual datasets.
A key conclusion of our study is that the problem of bias discovery and quantification in visual datasets is still open, and there is room for improvement in terms of both methods and the range of biases that can be addressed. Moreover, there is no such thing as a bias-free dataset, so scientists and practitioners must become aware of the biases in their datasets and make them explicit. To this end, we propose a checklist to spot different types of bias during visual dataset collection. 
\end{abstract}

\section{Introduction}
\label{Introduction}
In the fields of Artificial Intelligence (AI), algorithmic fairness, and (big) data ethics, the term \textit{bias} usually refers to the case in which AI-powered decisions show prejudice against individuals or groups of people defined based on protected attributes like gender or race \citep{Ntoutsi2020}. Instances of this prejudice have caused discrimination in many fields, including recidivism scoring \citep{Angwin2016}, online advertisement \citep{Sweeney13}, facial recognition \citep{Cook2019}, and credit scoring \citep{Bartlett2019}. 

Defining the concepts of bias and fairness in mathematical terms is not a trivial task. \cite{Verma2018} provided a survey on more than 20 different measures of algorithmic fairness, many of which are incompatible with each other.
This incompatibility - the so-called impossibility theorem \citep{Chouldechova2017, Kleinberg17} - forces scientists and practitioners to choose the measures they use based on their personal beliefs or other constraints (e.g., business models) on what has to be considered fair for the particular problem/domain.

While algorithms may also be responsible for the amplification of pre-existing biases in the training data \citep{Bolukbasi2016}, the quality of the data itself contributes significantly to the development of discriminatory AI applications. \cite{Ntoutsi2020} identified two ways in which bias is encoded in the data: correlations and causal influences among the protected attributes and other features; and the lack of representation of protected groups in the data. They also noted that biases manifest in ways that are specific to the data type. 

In this work, we focused on how biases can be encoded in the data (e.g., via spurious correlations, causal relationship among the variables, and unrepresentative data samples) and, in particular, in \emph{visual} data (i.e., images and videos), which comprises one of the most popular and complex data types. Visual data encapsulates many features that require human experience and context to interpret. These include the human subjects, how they are depicted and their reciprocal position in the image frame, implicit references to culture-specific notions and background knowledge, etc. Even the colouring scheme can convey different messages. Thus, making sense of visual content remains a very complex task, and understanding bias in visual data is even harder.

Computer vision (CV), the primary domain that enables computers to gain high-level understanding from visual data, is heavily dominated nowadays by deep learning (DL) methods \citep{LeCun15, Baraniuk2020} that allowed for outstanding performance in tasks like object detection, image classification and image segmentation. DL methods, however, rely heavily on data, and the results are as good and ``fair'' as the data used for their training. CV has recently drawn attention for its ethical implications when deployed in several settings, ranging from targeted advertising to law enforcement. There has been mounting evidence that deploying CV systems without a comprehensive ethical assessment may result in major discrimination against protected groups. For instance, facial recognition technologies \citep{Cook2019, Robinson2020}, gender classification algorithms \citep{Buolamwini2018}, and autonomous driving systems \citep{Wilson2019} have been all shown to exhibit discriminatory behaviour. 

While bias in AI systems is a well-studied field, the research in biased CV is more limited despite the abundance of visual data produced nowadays and their widespread use in the ML community. Moreover, to the best of our knowledge, there is no comprehensive survey on bias in visual datasets (\citep{Torralba2011} represents a seminal work in the field, but it is limited to object detection datasets). Hence, the contributions of the present work are: i) to explore and discuss the different types of bias that arise from the collection of visual data; ii) to systematically review the works that aim at addressing and measuring bias in visual datasets; iii) to discuss some attempts to compile bias-aware datasets. We believe this work to be a useful tool for helping scientists and practitioners to both develop new bias-discovery methods and collect data in ways as less biased as possible. To the latter end, we propose a checklist that can be used to spot the different types of bias that might enter the data during the collection process (Table \ref{tab:checklist}). 

The structure of the survey is as follows.
First, we describe in detail the different types of bias that might affect visual datasets (Section \ref{sec:CVBias}), provide concrete examples of CV applications that are affected by those biases, and a description of how they manifest in the life cycle of visual content. Second, we systematically review the methods for bias discovery in visual content proposed in the literature (Section \ref{sec:Discovery}). Third, in Section \ref{sec:BiasFree}, we discuss the weaknesses and strengths of some bias-aware visual benchmark datasets. Finally, in Section \ref{sec:Conclusions}, we conclude and outline some possible future research direction.

\section{Manifestation of Bias in Visual Data}
\label{sec:CVBias}

\begin{figure*}
    \centering
    \includegraphics[width = 430pt]{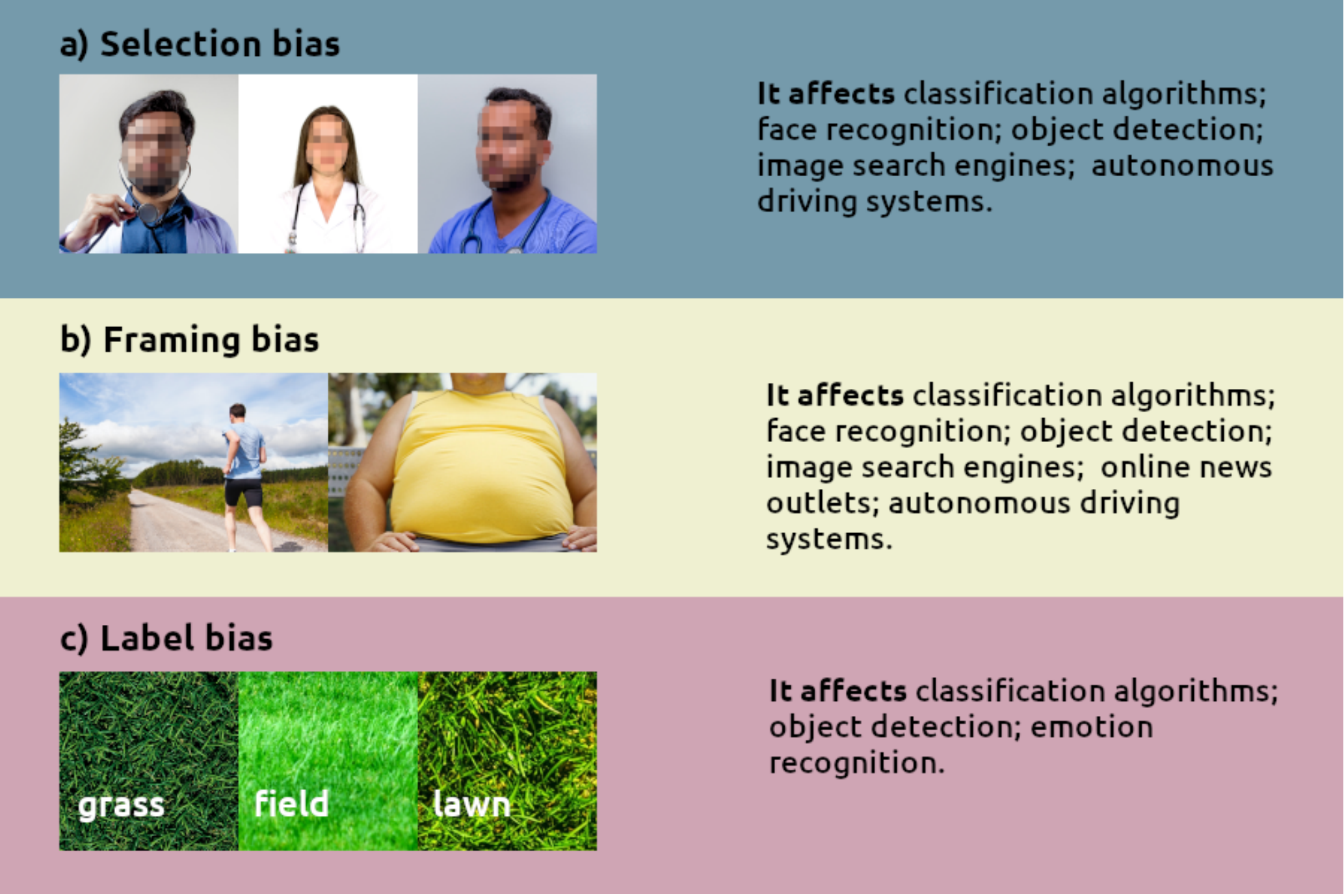}
    \caption{Examples of selection, framing and label bias. On the right, a list of applications that can be affected by each type of bias.}
   \label{fig:biases}
\end{figure*}

In this section, we describe in detail the types of bias that pertain to the \emph{capture} and \emph{collection} of visual data (Figure \ref{fig:biases}), namely: selection bias (Section~\ref{sec:selectionbias}), framing bias (Section~\ref{sec:framingbias}) and label bias (Section~\ref{sec:labelbias}). Furthermore, we describe (Section \ref{subsec:LifeCycle}) how they manifest within the life cycle of visual content, from capture to the deployment of CV algorithms. Note that a comprehensive analysis of historical discrimination and algorithmic bias is beyond the scope of this work. The interested reader can refer to \cite{Bandy2021} for a survey on methods for auditing algorithmic bias both in CV and other AI-related areas.

Our categorisation builds on the scheme by \cite{Torralba2011} who organised the types of visual bias into four different categories: \emph{selection bias}, \emph{capture bias} (which we collapse into the more general concept of \emph{framing bias}), \emph{label bias}, and \emph{negative set bias}. The latter arises when the labelling does not reflect entirely the population of the negative class (say \emph{non-white} in a binary feature [white people/non-white people]). We consider negative class bias as an instance of selection and label bias. 

Even though our categorisation appears on the surface to be similar to the one by \cite{Torralba2011}, their analysis focused on datasets for object detection. Instead, we contextualise bias in a more general setting and we also focus on discrimination against protected groups\footnote{Note that, while this is the focus of our survey, we also take into account cases in which bias does not necessarily affect people, e.g., in object detection.}. Since selection, framing and label bias manifest in many different ways, we also go further by describing a sub-categorisation of these three types of bias (Table \ref{tab:sub_biases}) including several biases commonly encountered in Statistics, Health studies, or Psychology and adapting them to the context of visual data. While in the following we describe selection, framing, and label bias in general terms, we also provide references in Table \ref{tab:sub_biases} for the interested reader who might want to delve further into their different manifestations.

\subsection{Selection Bias} 
\label{sec:selectionbias}

\paragraph{Definition} Selection bias is the type of bias that ``occurs when individuals or groups in a study differ systematically from the population of interest leading to a systematic error in an association or outcome"\footnote{Catalogue of Bias Collaboration, Nunan D, Bankhead C, Aronson JK. Selection bias. Catalogue Of Bias 2017: \url{http://www.catalogofbias.org/biases/selection-bias/}. Last visited 18.05.2022.}. More generally, it refers to any ``association created as a result of the process by which individuals are selected into the analysis" (\cite[Chapter 8, pg. 99]{CausalInferenceWhatIf}). In visual datasets, using the first definition would be tricky as, for instance, in the case of facial recognition, respecting the ethnic composition of the population is generally not enough to ensure good performance across every subgroup, as we will see in the following. Hence, we adopt a slight modification of \cite{CausalInferenceWhatIf} definition: \\

\noindent
\begin{minipage}[t]{0.45\textwidth}
    \fbox{%
    \parbox{\textwidth}{%
\emph{We call} selection bias \emph{any disparities or associations created as a result of the process by which subjects are included in a visual dataset}. }}
\end{minipage}\\

\paragraph{Description} \cite{Torralba2011} showed that certain kinds of imagery are more likely to be selected during the collection of large-scale benchmark datasets, leading to selection bias (see \emph{sampling bias}, Table \ref{tab:sub_biases}). For example, in Caltech101 \citep{FeiFei04} pictures labelled as cars are usually taken from the side, while ImageNet \citep{Deng09} contains more racing cars. Furthermore, a strong selection bias can also be present within datasets. Indeed, \cite{Salakhutdinov2011} showed that, unless a great effort is made to keep the distribution uniform during the data collection process, categories in large image datasets follow a long-tail distribution. Having such a distribution means that, for example, people and windows are way more common than ziggurats and coffins \citep{Zhu2014}. 

Selection bias becomes particularly worrisome when it concerns humans. \cite{Buolamwini2018} pointed out that under-representation of people from different genders and ethnic groups may result in a systematic misclassification of these groups (their work concentrates on gender classification algorithms). They also showed that some popular datasets were biased towards lighter-skinned male subjects. For example, Adience \citep{Eidinger14} (resp. IJB-A \citep{Klare15}) 
have 7.4\% (resp. 4.4\%) of darker-skinned females, 6.4\% (resp. 16.0\%) of darker-skinned males, 44.6\% (resp. 20.2\%) of lighter-skinned females and 41.6\%  (resp. 59.4\%) of lighter-skinned males. Such imbalances greatly affect the performance of CV tools. For instance, \cite{Buolamwini2018} showed that the error rate for dark skin individuals could be 18 times higher than for light skin ones in some commercial gender classification algorithms. 

\paragraph{Affected applications} In summer 2020, the New York Times published the story of a Black American individual wrongfully arrested due to an error made by a facial recognition algorithm \citep{NYTimesJune2020}. While we do not know whether bias in the data caused this exact case, we know that selection bias can lead to different error rates in face recognition. Hence, such technology would require much more care, especially in high-impact applications. 

Autonomous driving systems are also likely affected by selection bias as it is very challenging to collect a dataset that describes every possible scene and situation a car might face. The Berkeley Driving Dataset \citep{Yu2020} for example, contains driving scenes from only four cities in the US; an autonomous car trained on such a dataset will likely under-perform in other cities with different visual characteristics. The effect of selection bias on autonomous driving becomes particularly risky when it affects pedestrian recognition algorithms. \cite{Wilson2019} studied the impact of under-representation of darker-skinned people on the predictive inequity of pedestrian recognition systems. They found evidence that the effect of this selection bias is two-fold: first, such imbalances ``beget less statistical certainty" making the process of recognition more difficult; second, standard loss functions tend to prioritise the more represented groups and hence some kind of mitigation measures are needed in the training procedure \citep{Wilson2019}.

Moreover, \cite{Buolamwini2018} explained that part of the collection of benchmark face datasets is often done using facial detection algorithms. Therefore, every systematic bias in the training of those tools propagates to other datasets. This is a clear example of \textit{algorithmic bias turning into selection bias} (see \emph{automation bias}, Table \ref{tab:sub_biases}), as described at the end of Section \ref{subsec:LifeCycle}. Furthermore, image search engines can contribute to the creation of selection biases (see \emph{availability bias}, Table \ref{tab:sub_biases}) as well due to systematic biases in their retrieval algorithms; see, \citep{Kay2015} for a study on gender bias in Google's image search engine. 

\paragraph{Remarks} Finally, \cite{Klare2012} pointed out that, while class imbalances have undoubtedly significant impact on some facial recognition algorithms, they do not explain every disparity in the performance of algorithms. For instance, they suggested that a group of subjects might be more challenging to recognise, even with balanced training data, if it is associated with higher variance (for example, due to hairstyle or make-up). The reader can also refer to \cite{Terhoerst2021a} for an analysis of the effect of these kinds of attributes on facial recognition technology.

\subsection{Framing Bias}
\label{sec:framingbias}

\paragraph{Definition} According to the seminal work of \cite{Entman93} on framing of (textual) communication, ``To frame is to \emph{select some aspects of a perceived reality and make them more salient in a communicating text, in such a way as to promote a particular problem definition, causal interpretation, moral evaluation and/or treatment recommendation} for the item described". \cite{Coleman2010} adopted the same definition for the framing of visual content and added that ``In visual studies, \emph{framing} refers to the selection of one view, scene, or angle when making the image, cropping, editing or selecting it". These two definitions highlight how framing bias has two different aspects. First, the medium aspect: visual content is, in all respect, a medium, and therefore the way they are composed conveys different messages. Second, the technical aspect: framing bias also derives from how an image is captured or edited. Hence, in the following\\

\noindent
\begin{minipage}[t]{0.45\textwidth}
    \fbox{%
    \parbox{\textwidth}{%
 \emph{We refer to \emph{framing bias} as any associations or disparities that can be used to convey different messages and/or that can be traced back to the way in which the visual content has been composed}.}}
\end{minipage} \\

Note that, while the selection process is a powerful tool for framing visual content, we keep the concepts of selection and framing bias distinct, as they present their own peculiarities.

\paragraph{Description} An example of visual framing bias (to be more specific, \emph{capture bias}, Table \ref{tab:sub_biases}) has been studied by \cite{Heuer2011}. They analysed images attached to obesity-related articles that appeared on major US online news websites in the period 2002-2009. They concluded that there was a substantial difference in how such images depicted obese people with respect to non-overweight ones. For example, 59\% of obese people were headless (6\% of non-overweight), and  52\% had only the abdomen portrayed (0\% of non-overweight). This portrayal as ``headless stomachs" \citep{Heuer2011} (see Figure \ref{fig:biases} b) for an example) may have a stigmatising and de-humanising effect on the viewer. On the contrary, positive characteristics were more commonly portrayed among non-overweight people. \cite{Corradi2012}, while analysing the use of female bodies in advertisements, talked about ``semiotic mutilation" when parts of a woman's body are used for advertising a product with a de-humanising effect, similar to what Heuer et al. have described in their work about obesity.

The relationship between bodies and faces in image framing is a well-known problem. Indeed, \cite{Archer1983} used a ratio between the height of a subject's body and the length of their face to determine whether men's faces were more prominent than those of women. They found out that this was true in three different settings: contemporary American news media photographs, contemporary international news media photographs, and portraits and self-portraits from the 17-th century to the 20-th century (interestingly, there was no disparity in earlier artworks). Furthermore, they found some evidence that people tend to draw men with higher facial prominence and thus that this bias does not only occur in mass media or art. 

\paragraph{Affected applications} An application that can suffer greatly from framing bias (see \emph{stereotyping}, Table \ref{tab:sub_biases}) is that of image search engines. For example, \cite{Kay2015} found out that while searching for construction workers on Google Image Search, women were more likely to be depicted in an unprofessional or hyper-sexualised way\footnote{This specific problem seems to be solved, but the queries \texttt{female construction worker} and \texttt{male construction worker} still return several pictures in which the subjects are hyper-sexualised.}. However, it is unclear whether the retrieval algorithms are responsible for the framing or they just index popular pages associated with the queries. Nonetheless, the problem remains because the incorrect framing in a search engine can contribute to the spread of biased messages. We recall, for instance, the case of the photo of a woman ``ostensibly about to remove a man's head" \citep{WPApril2015} that was retrieved after searching ``feminism" on a famous stock photos website.

\subsection{Label Bias}
\label{sec:labelbias}

\paragraph{Definition} For supervised learning, labelled data are required. The quality of the labels\footnote{Note that by label we mean any tabular information attached to the image data (object classes, measures, protected attributes, etc.)} is of paramount importance for learning and comprises a tedious task due to the complexity and volume of today's datasets. \cite{Jiang2020} define label bias as ``\emph{the bias that arises when the labels differ systematically from the ground truth}''. For instance, \cite{Terhoerst2021} found some evidence that the face recognition dataset Labelled Faces in the Wild (LFW, see \citep{Huang08} for the original release and \citep{Kumar09} for the attributes) presents a very low label accuracy against human annotation. 

Furthermore, \cite{Torralba2011} highlight bias as a result of the labelling process itself for reasons like ``semantic categories are often poorly defined, and different annotators may assign different labels to the same type of object". Torralba and Efros' work mainly focused on object detection tasks, and hence by \emph{different label assignment} they refer, e.g., to ``grass'' labelled as ``lawn'' or ``picture'' as ``painting''. Nevertheless, these problems arise especially when dealing with human-related features such as race or gender.\\

\noindent
\begin{minipage}[t]{0.45\textwidth}
    \fbox{%
    \parbox{\textwidth}{%
\emph{We define \emph{label bias} as any errors in the labelling of visual data, with respect to some ground truth, or the use of poorly defined or inappropriate semantic categories}.}}
\end{minipage} \\

\paragraph{Description} As already mentioned, a significant source of labelling bias is the \emph{poor definition of semantic categories}. Race is a particularly clear example of this: according to \cite{Barbujani2011} ``The obvious biological differences among humans allow one to make educated guesses about an unknown person’s ancestry, but agreeing on a catalogue of human races has so far proved impossible". Given such an impossibility, racial categorisation in visual datasets must come at best from subjects' own race perception or, even worse, from the stereotypical bias of annotators. From a CV standpoint then, it would be probably more accurate to use actual visual attributes, if strictly necessary, such as \emph{Fitzpatrick skin type} \citep{Buolamwini2018} or \emph{skin reflectance} \citep{Cook2019} rather than a fuzzy category such as race. Note that, while skin tone can be a more objective trait, it still does not entirely reflect human diversity.
Similarly, the binary categorisation of gender has been criticised. The reader can refer to \citep{Hanna20, Kasirzadeh21} to explore the challenges that the use of ill-defined categories poses to algorithmic fairness from both the ontological and operational points of view.

Moreover, \cite{Torralba2011} argued that different annotators can come up with different labels for the same object. While this mainly applies to the labelling of object detection datasets rather than face datasets, where labels are usually binary or discrete, it gives us an important input about bias in CV in general: annotators' biases and preconceptions are reflected in the datasets.

We can view what has been described so far also as a problem of operationalisation of what \cite{Jacobs21} called unobservable theoretical constructs (see \emph{measurement bias}, Table \ref{tab:sub_biases}). They proposed a framework that serves as guideline for the mindful use of fuzzy semantic categories and answers the following questions on the validity of  operationalisations (or measurements) of a construct: ``Does the operationalization capture all relevant aspects of the construct purported to be measured? Do the measurements look plausible? Do they correlate with other measurements of the same construct? Or do they vary in ways that suggest that the operationalization may be inadvertently capturing aspects of other constructs? Are the measurements predictive of measurements of any relevant observable properties (and other unobservable theoretical constructs) thought to be related to the construct, but not incorporated into the operationalization? Do the measurements support known hypotheses about the construct? What are the consequences of using the measurements[?]" 

A concrete case of the use of a fuzzy semantic category is provided in \citep{Liang2018}, where the creation of a face dataset for facial beauty is described. Since beauty and attractiveness are prototypical examples of subjective characteristics, it is obvious that attempts of constructing such datasets will be filled with the personal preconceptions of the participants who labelled the images (see \emph{observer bias}, Table \ref{tab:sub_biases}).

\paragraph{Affected applications} Since deep learning boosted the popularity of CV, a modern form of physiognomy has gained a certain momentum. Recently, several studies appeared claiming to be able to classify images according to the criminal attitude of the subjects\footnote{See, for instance, the retracted article: Hashemi and Hall \url{https://journalofbigdata.springeropen.com/articles/10.1186/s40537-019-0282-4}. Last visited 30.03.2022.} or sexual orientation\footnote{For example, Kosinski and Wang  \url{https://www.gsb.stanford.edu/faculty-research/publications/deep-neural-networks-are-more-accurate-humans-detecting-sexual}. Last visited 30.03.2022.}. A commercial tool has also been released to detect terrorists and paedophiles. While ``doomed to fail" for a series of technical reasons well explained by \cite{Bowyer2020}, these applications rely on a precise ideology that \cite{Goldenfein19} called \emph{computational empiricism}: an epistemological paradigm that claims, despite any scientific evidence, that the true nature of humans can be measured and unveiled by algorithms. The reader can also refer to the famous blog post ``Physiognomy's New Clothes"\footnote{B. Ag\"uera y Arcas, M. Mitchell and A. Todorov, \textit{Physiognomy’s New Clothes}, May 2017. \url{https://medium.com/@blaisea/physiognomys-new-clothes-f2d4b59fdd6a}. Last visited 30.03.2022.} for an introduction to the problem. 

\paragraph{Remarks} Just as selection bias, label bias can lead to a vicious cycle: a classification algorithm trained on biased labels will most likely reinforce the original bias when used to label newly collected data (see \emph{automation bias}, Table \ref{tab:sub_biases}). 

\input{table_sub_biases}

\subsection{Media Bias Life Cycle}
\label{subsec:LifeCycle}

\begin{figure*}
    \centering
    \includegraphics[width =430pt]{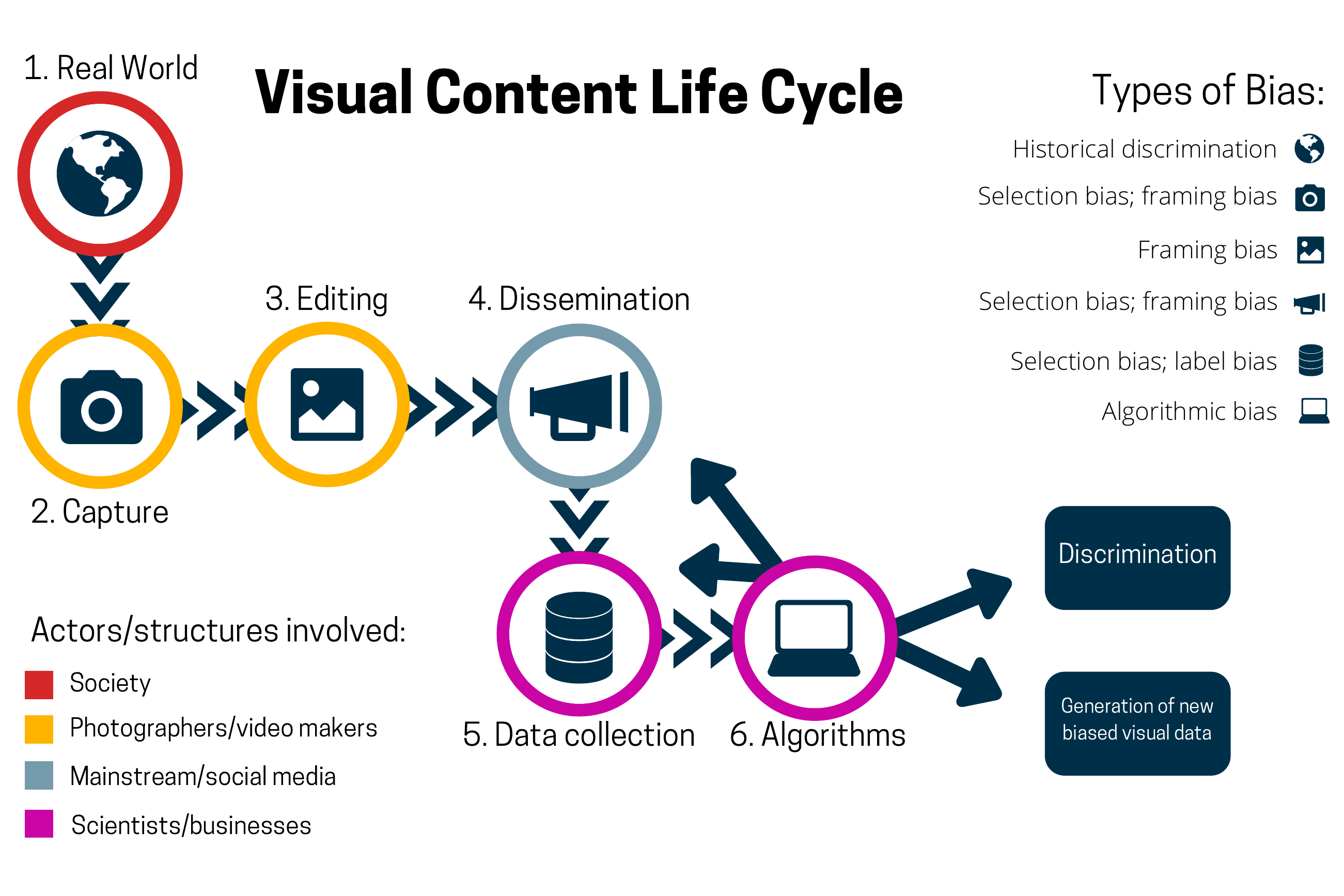}
    \caption{Simplified illustration of visual content life cycle and associated sources of bias.}
    \label{fig:LifeCycle}
\end{figure*}

Figure~\ref{fig:LifeCycle} gives an overview of the life cycle of visual content and depicts how different types of bias can enter at any step of this cycle and can be amplified in consecutive interactions. Below we describe each of these components in more detail.

\paragraph{1. Real world} The journey of visual content alongside bias starts even before the content is generated. Inequalities undeniably shape our world, and this is reflected in the generation of data in general and of visual content in particular. For example, \cite{Zhao2017} found out that the dataset MS-COCO \citep{Lin2014}, a large-scale object detection, segmentation, and captioning dataset which is used as a benchmark in CV competitions, was more likely to associate kitchen objects with women. While both image capturing and dataset collection come at a later stage in the life cycle described in Figure \ref{fig:LifeCycle}, it is clear that in this instance, such bias has roots in the gender division between productive and reproductive/care labour.
Nevertheless, as shown in the following paragraphs, each step of the life cycle of visual content can reproduce or amplify historical discrimination as well as insert new biases. 
 
\paragraph{2. Capture} The actual life of visual content starts with its capture. Here the first types of bias can be introduced, \emph{selection bias} and \emph{framing bias}. While selection bias is usually observed in datasets, where entire groups can be under-represented or non-represented at all, the selection begins with the choices of the photographer/video maker\footnote{We do not necessarily mean \emph{professional} photographers/video makers.}. Moreover, the way a photo is composed (arrangement, camera's setting, lighting conditions, etc.) is a powerful way of conveying different messages and thus a possible source of framing bias. Note that the selection of the subjects and the framing of the photo/video are both active choices of the photographer/video maker. Nevertheless, historical discrimination can turn into selection and framing bias if not actively countered.

\paragraph{3. Editing} With the advent of digital photography, image editing and post-processing are now a key step in the content life cycle. Post-processing has become a fundamental skill for every photographer/video maker, along with skills such as camera configuration, lighting, shooting, etc. Since photo editing tools are extremely powerful, they could give rise to many ethical issues: to what extent and in which contexts is it right to modify the visual content of an image or video digitally? 
What harms could such modifications potentially cause?
The discussion around the award-winning Paul Hansen's photo ``Gaza Burial" \citep{DerSpiegelMay2013} represents a practical example of how these questions are important to photojournalism. In particular, what is the trade-off between effective storytelling and adherence to reality? Nonetheless, photo editing does not concern journalism only. Actually, it affects people, and especially women, in several different contexts, from the fashion industry \citep{BBCDecember2018} to advertising and high-school yearbooks \citep{NYTimesMay2021}. 

\paragraph{4. Dissemination} The next step in the life of visual content is dissemination. Nowadays, images and videos are shared via social and mainstream media in volumes that nobody can possibly inspect. For instance, more than 500 hours of videos are uploaded to YouTube every minute\footnote{\url{https://blog.youtube/press/}. Last visited 23.03.2022.}. Dissemination of visual content suffers from both selection and framing bias. The images that are selected, the medium and channels through which they are disseminated, the caption or text that are attached to them are all elements that could give rise to bias (see \citep{Peng2018} for a case study of framing bias in the media coverage of 2016 US presidential election). 

\paragraph{5. Data collection} The part of the life cycle of visual content that is more relevant to our discussion is the collection of visual datasets. Here we encounter selection bias once again, as the data collection process can exclude or under-represent certain groups from appearing in the dataset, as well as \emph{label bias} as great effort is usually expended to collect data along with additional information in the form of annotations. As we discussed in Section \ref{sec:labelbias}, this process is prone to errors, mislabelling, and explicit discrimination (see \citep{Miceli20} for an analysis of power dynamics in the labelling process of visual data). Furthermore, benchmark datasets have gained incredible importance in the CV field. On the one hand, they allowed significant measurable progress in the field. On the other hand, they represent only a small portion of the visual world (see \citep{Prabhu2020} for a critique of the large-scale collection of visual data \emph{in the wild}). 

\paragraph{6. Algorithms} Finally, there is the actual step of model training. Fairness and accountability of algorithms is a pressing issue as algorithm-powered systems are used pervasively in applications and services impacting a growing number of citizens \citep{Drozdowski2020, Bandy2021}. Important questions arise for the AI and CV communities on how to implement fairness in algorithms such as: What legal frameworks should governments put into place? What are the strategies to mitigate bias or make algorithms explainable? 
Nevertheless, the journey of visual content does not end with the training of models. 
Indeed, there are several ways in which biased algorithms can generate vicious feedback loops as described in the remarks of Section \ref{sec:labelbias}.
Moreover, the recent explosion in popularity of generative models, namely Generative Adversarial Networks (GANs, \citep{Goodfellow14}), has made the process of media creation very easy and fast (see \citep{Mirsky2021} for a survey on AI-generated visual content). Such AI-generated media can then be reinserted in the content life cycle via the Web and present their own ethical issues (see \citep{DBLP:journals/dagstuhl-manifestos/BerendtGHHHKNS21} for a discussion on the socio-technical aspects and ambivalence of the Web of humans and AIs). 

\section{Bias Discovery and Quantification in Visual Datasets}

\label{sec:Discovery}
\input{table_notation}

This section aims to understand how researchers have tackled the problem of discovery and quantification of bias in visual datasets since high-quality visual datasets are a critical ingredient towards fair and trustworthy CV systems \citep{Hu2020}. 
To this end, we performed a systematic survey of papers addressing the following problem: \textit{Given a visual dataset $\mathcal{D}$, is it possible to discover/quantify what types of bias it exhibits (of those described in Section \ref{sec:CVBias})?} In particular, we focus on the methodologies and measures used in the bias discovery process, of which we are going to outline the pros and cons. Furthermore, we will try to define open issues and possible future directions for research in this field. Note that this problem is critically different from both the problem of finding out whether an algorithm discriminates against a protected group and the problem of mitigating such bias. 

In order to systematise this review, we proceed in the following way to collect the relevant material: 
first, we outlined a set of keywords to be used in three different scientific databases (DBLP, arXiv and Google Scholar) 
and we selected only the material relevant to our research question following a protocol described in the following paragraph; second, we summarised the results of our review and outlined the pros and cons of different methods in Section \ref{sec:prosCons}. Our review methodology was inspired by the works of \cite{Merli2018}, and \cite{Kofod2012}. Our protocol also resembles the one described in \cite[Section 5.1.1]{kitchenham04}.

Given our problem, we identify the following relevant keywords: \emph{bias}, \emph{image}, \emph{dataset}, \emph{fairness}. For each of them, we also defined a set of synonyms and antonyms (see Table \ref{tab:keywords}). Note that we include the words ``face" and ``facial" among the synonyms of the word image. This is mainly motivated by the fact that in the title and the abstract of the influential work of \cite{Buolamwini2018} there are no occurrences of the word ``image". Instead, we find many occurrences of the expression ``facial analysis dataset". Since facial analysis is an important case study for detecting bias in visual content, it makes sense to include it explicitly in our search. The search queries have been composed of all possible combinations of the different synonyms (antonyms) of the above keywords (for example, ``image dataset bias", ``visual dataset discrimination", or ``image collection fairness"). 

\begin{table*}
\centering
\begin{tabular}{ |l|l|l| }
\cline{2-3}
\multicolumn{1}{c|}{} & \textbf{Synonyms} & \textbf{Antonyms} \\
\hline
 \textbf{bias} & discrimination & fair, unbiased \\ 
 \textbf{image} & visual, face, facial &  \\  
 \textbf{dataset} & collection &   \\
 \hline
\end{tabular}
\caption{Set of keywords and relative synonyms and antonyms.}
\label{tab:keywords}
\end{table*}

This process resulted, after manual filtering\footnote{The manual filtering consisted in keeping only those papers that describe a method or a measure for discovering and quantifying bias in visual datasets. In some cases, we kept also works developing methods for algorithmic bias mitigation but that could be used to discover bias in the dataset as well (see \cite{Balakrishnan2020}).}, in 17  relevant papers.
We expanded the list with 6 articles by looking at papers citing the retrieved works (via Google Scholar) and their ``related work" sections. We also added to the list the work of \cite{Lopez-Paz2017} on causal signals in images that was not retrieved by the protocol described above.
In the following, we review all these 24 papers, dividing them into four categories according to the strategies they use to discover bias: 

\begin{itemize}

\item \textbf{Reduction to tabular data:} these rely on the attributes and labels attached to or extracted from the visual data and try to measure bias as if it were a tabular dataset.

\item \textbf{Biased image representations:} these rely on lower-dimensional representations of the data to discover bias.

\item  \textbf{Cross-dataset bias detection:} these  assess bias by comparing different datasets, trying to discover some sort of ``signature'' due to the data collection process.

\item \textbf{Other methods:} Different methods that could not fit any of the above categories.

\end{itemize}

\input{table_metrics}

\subsection{Reduction to Tabular Data} 
\label{subsec:Reduction}
Methods in this category transform visual data into tabular format and leverage the multitude of bias detection techniques developed for tabular datasets~\citep{Ntoutsi2020}. The features for the tabular description can be extracted either \emph{directly} from the images (using, for example, some image recognition tools) or \emph{indirectly} from some accompanying image description/annotation (e.g., image captions, labels/hashtags) or both. In the majority of cases, feature extraction relies on automatic processes and is therefore prone to errors and biases. Thus, biases that exist in the original images (selection, framing, or label bias) might be reflected and even amplified in the tabular representation due to the bias-prone feature extraction process.

Below we review different approaches under this category that extract protected attributes from visual datasets and evaluate whether bias exists. Existing approaches can be broadly categorized into parity-based, information theoretical and others. As already explained, the sources of bias in such a case are not limited to bias in the original images, but bias may also exist due to the labelling process and the automatic feature extraction. The impact of such additional sources on the results is typically omitted.

\paragraph{Count/demographic parity} 

\cite{Dulhanty2019} proposed a simple method for auditing ImageNet \citep{Deng09} with respect to gender and age biases. They applied a face detection algorithm to two different subsets of ImageNet, the training set of ILSVRC \citep{Russakovsky15} and the \textit{person} category of ImageNet \citep{Deng09}. After that, they applied an age recognition model and a gender recognition model to them. They then computed the dataset distribution across the age and gender categories, finding out a prevalence of men (58.48\%) and a minimal amount of people (1.71\%) in the $>60$ age group. Computing the percentage of men and women across every category allowed them to identify the highest class imbalances in the two subsets of ImageNet. 
For example, in the ILSVRC subset, the 89.33\% of the images in category \textit{bulletproof vest} were labelled as men and the 82.81\% of the images in category \textit{lipstick} were labelled as women. Therefore, this method does not only give information on the selection bias but also on the \emph{framing} of the protected attributes, given a suitable labelling of the dataset. The authors noted that this method relies on the assumption that the gender and age recognition models involved are not biased. As the authors pointed out, such an assumption is violated by the gender recognition model, and therefore, analysis cannot be totally reliable. 

\cite{Yang2020} performed another analysis of the \textit{person} category of ImageNet \citep{Deng09}, trying to address both selection and label bias. They addressed label bias by asking annotators to find out, first, whether the labels could be offensive or sensitive (e.g., sexual/racial slur), and, second, to point out whether some of the labels were not referring to visual categories (e.g., it is difficult to understand whether an image depicts a \textit{philanthropist}). They removed such categories and continued their analysis by asking annotators to further label images according to some categories of interest (gender, age, and skin colour) to understand whether the remaining data were balanced with respect to those categories and then to address selection bias. This demographic analysis showed that women and dark-skinned people were both under-represented in the remaining non-offensive/sensitive and visual categories. Moreover, despite the overall under-representation, some categories were found to align with stereotypes (e.g., the 66.4\% of people in the category \emph{rapper} were dark-skinned). Hence, they also potentially addressed some framing bias. The annotation process was validated by measuring the annotators' agreement on a small controlled set of images. 

\cite{Zhao2017} measured the correlation between a protected attribute and the occurrences of various objects/actions.
They assumed to have a dataset labelled with a protected attribute $G$ with values $\{g_1,...,g_n\}$ and an attribute $O=o$ describing the occurrences of said objects or actions in the images (for instance, $G$ could be the gender and $O$ a cooking tool or an outdoor sport scene). The bias score of the dataset with respect to a protected attribute value $G = g$ and the scene/object $O=o$ is defined as the percentage of $g$ co-occurrences with $o$: 
\begin{equation}
    b(o,g) = \frac{c(o,g)}{\sum_{x\in \{g_1,...,g_n\}} c(o,x)}
\end{equation}
where $c(o,x)$ counts the co-occurrences of the object/scene $o$ and the protected attribute's value $x$. If $b(o,g_i)>\frac{1}{n}$, where $n$ is the number of values that the protected attribute can assume, it means that attribute $G=g$ is positively correlated with the object/action $O=o$.

\cite{Shankar2017} used instead a simple count to assess geographic bias in ImageNet \citep{Russakovsky15} and Open Images \citep{Krasin16}. They found out, for example, that the great majority of images of which the geographic provenance is known comes from the USA or Western European countries, resulting in highly imbalanced data. Such a geography-based analysis has been expanded by \cite{Wang2020}. While having balanced data is important for many applications, a mere count is usually not enough to assess every type of bias, as a balanced dataset could still contain spurious correlations, framing bias, label bias, etc. 

\cite{Buolamwini2018} constructed a benchmark dataset for gender classification. For testing discrimination in gender classification models, their dataset is balanced according to the distribution of both gender and Fitzpatrick skin type as they noticed that the error rates of classification models tended to be higher at the intersection of those categories (e.g., black women) because of the use of imbalanced training data. Hence, while they quantify bias by simply counting the instances with certain protected attributes, the novelty of their work is that they took into account multiple protected attributes at a time.

\paragraph{Information theoretical}
 
\cite{Merler2019} introduced four measures to construct a balanced dataset of faces. Two measures of diversity: Shannon entropy $H(X)=-\sum_{i=1}^n\mathbb{P}(X=x_i)\cdot ln(\mathbb{P}(X=x_i))$ and Simpson index $D(X)=\frac{1}{\sum_{i=1}^n\mathbb{P}(X=x_i)^2}$ where $\mathbb{P}(X=x_i)$ is the probability of an image having value $x_i$ for the attribute $X\in\{x_i\}_{i=1}^n$; and two measures of evenness: ($E_{Shannon}=\frac{H(X)}{ln(n)}$ and $E_{Simpson}=\frac{D(X)}{n}$). Such measures have been applied to a set of facial attributes, ranging from craniofacial distances to gender and skin colour, computed both via automated systems and with the help of human annotators.

\cite{Panda2018} also proposed to use (conditional) Shannon entropy for discovering \emph{framing bias} in emotion recognition datasets. Using a pre-trained model, they computed the dataset's top occurring objects/scenes. They computed the conditional entropy of each object across the positive and negative set of the emotions to see whether some objects/scenes were more likely to be related to a certain emotion. For example, they found out that objects like balloons or candy stores are only present in the negative-set of \emph{sadness} in Deep Emotion \citep{You2016}. Given an object $c$ and an emotion  $E=e\in\{0,1\}$ (where $e=1$ represents, for instance, \textit{sadness} and $e=0$ represents the negative-set \textit{non-sadness}) they computed: 

\begin{equation}
    H(E|X = c) = -\sum_{e\in\{0,1\}}p_{ec}\cdot ln(p_{ec}) 
\end{equation} 
where $$p_{ec} = \mathbb{P}(E=e|X=c)$$
When the conditional entropy of an object is zero, it means that such an object is associated only with the emotion $E$ or, on the contrary, is never associated with it (it only appears in the negative set). This may be considered a type of framing bias. 

\cite{Kim2019} introduced another definition of bias inspired by information theory. They wanted to develop a method for training classification models that do not overfit due to dataset biases. In doing so, they give a precise definition of bias: a dataset contains bias when $I(X, Y) \gg 0$ where $X$ is the protected attribute, $Y$ is the output variable and $I(X, Y):= H(X)-H(X|Y)$ is the mutual information between those random variables. Kim et al. proposed to minimise such mutual information during training so that the model forgets the biases and generalises well. Note that if $Y$ is any feature in the dataset, this measure can be used to quantify bias in the dataset instead of the output of a model. 

\paragraph{Other}
\cite{Wang2019} defined \textit{dataset leakage} to measure the possibility of a protected attribute to be retrieved using only the information about non-protected ones. Given a dataset $\mathcal{D}=\{(x_i,y_i,g_i)\}$ where $x_i$ is an image, $y_i$ a non-protected attribute and $g_i$ is a protected attribute, the attribute $y_i$ is said to leak information about $g_i$ if there exists a function $f(\cdot)$, called \textit{attacker}, such that $f(y_i)\approx g_i$. Operationally, the attacker $f(\cdot)$ is a classifier trained on $\{y_i, g_i\}$. The dataset leakage is measured as follows: 
\begin{equation}
    \lambda_\mathcal{D}=\frac{1}{|\mathcal{D}|}\sum_{i=1}^{|\mathcal{D}|}\mathbf{1}[f(y_i)=g_i]
\end{equation}

\cite{Wachinger2021} explicitly used causality for studying spurious correlations in neuroimaging datasets. Given variables $X$ and $Y$ they wanted to test whether $X$ causes $Y$ or there exists a confounder variable $Z$ instead. Since those two hypotheses imply two different factorisations of the distribution $\mathbb{P}(X|Y)$, the factorisation with a lower Kolmogorov complexity is the one that identifies the true causal structure. Kolmogorov complexity is approximated by Minimum Description Length (MDL). 

\cite{Jang2019} proposed 8 different measures for identifying gender framing bias in movies. The following measures are computed for a movie for each gender, therefore the means are computed across every frame in which an actor of a certain gender appears. The measures are the following: Emotional Diversity $H(X) = -\sum_{i=1}^s\mathbb{P}(X=x_i)\cdot ln(\mathbb{P}(X=x_i))$ where $\mathbb{P}(X=x_i)$ is the probability that a character expresses a certain emotion $x_i$ and the sum is computed on the range of the different emotions shown by characters of a certain gender (the list of emotions was: anger, disgust, contempt, fear, happiness, neutral, sadness, and surprise); Spatial Staticity exploits ideas from time-series analysis to measure how static a character is during the movie, it is defined as $mean(PSD(p(t))$ where PSD is the power spectral density of the time-series of the position on the x-axis (resp. y-axis) of the character (the higher the value the less animated is the character); Spatial occupancy $mean(\sqrt{A})$ where $A$ is the area of the face of the character; Temporal occupancy $\frac{N}{N_{total}}$ where $N$ is the number of frames in which the character appears and $N_{total}$ is the total number of frames; Mean age computed over each frame and character; Intellectual Image $mean(G)$ where $G$ is the presence of glasses (this seems a debatable choice as it might itself suffer from some label bias); Emphasis on appearance $mean(E)$ where $E$ is the light exposure of faces again calculated for each frame; finally, the type and frequency of surrounding objects is analysed. The attributes involved in the computation of these measures are mainly extracted using Microsoft Face API, which \cite{Buolamwini2018} demonstrated to be biased against black women. 

\cite{Wang2020} proposed REVISE, a comprehensive tool for bias discovery in image datasets. The authors defined three sets of metrics. The first set contains metrics based solely on the use of bounding boxes of objects (Note that a \textit{person} is considered an \textit{object} as it can be classified by an objected detection model); the second set contains metrics that exploit also protected attribute labels; the third set uses additional unstructured information such as text when the protected attribute is not provided explicitly. REVISE implements 13 different metrics, some of which are very similar to what we described in the previous paragraphs. We are going to describe two of the most relevant: i) scene diversity $H(S)$ where $S$ is the scene attribute computed applying a pre-trained scene recognition algorithm \citep{Zhou18}; and ii) appearance differences, computed by extracting features using a feature extractor of images with a same object/scene but a different protected attribute value and then training an SVM classifier on the extracted features to see whether it could learn different representations of subjects with different protected attribute values in the same context. 

\subsection{Biased Image Representations}
\label{subsec:LatentRep}

The following methods analyse the distances and geometric relations among images exploiting their representation in a lower-dimensional space to discover the presence of bias. 

\paragraph{Distance-based}

\cite{Karkkainen2021} proposed a simple measure of diversity for testing their face dataset. They studied the distribution of pairwise L1-distances calculated after embedding the images in a lower-dimensional space using a neural network pre-trained on a different benchmark face dataset. If such distribution is skewed towards high pairwise distances, it means that the data show high diversity. Nonetheless, such an analysis is heavily influenced by the embedding. For instance, faces in a ``white-oriented'' dataset were also well separated in the embedding space, probably because the neural network used for the embedding had also been trained using a similarly biased dataset. 

\cite{Steed21} developed a method for addressing human-like biases in the latent image representation of unsupervised generative models inspired by bias detection methods in Natural Language Processing \citep{Caliskan2017}. They discovered that the biases found in the latent space of two big models trained on ImageNet \citep{Russakovsky15} match with human-like biases. The authors measured bias by looking at associations between semantic concepts (for example, man-career and woman-family) by measuring the cosine similarity among vectors in the latent space computed by applying the models to controlled samples of images that resemble those visual concepts. More precisely, given a model $f$ that maps images into a vector space $\mathbb{R}^d$, two sets of images $J$ and $K$ (e.g., photos of men and women, respectively), and two sets $A$ and $B$ of images representing the concepts we want to measure the association with (e.g., photos of people at work and photos of people in familiar settings), we can measure the association of $J$ with $A$ and $K$ with $B$ in the following way:
\begin{equation}
    s(J,K,A,B) = \sum_{j\in J}s(j,A,B) - \sum_{k\in K}s(k,A,B)
\end{equation}
where
\begin{equation}
\begin{aligned}
s(w,A,B) = & mean_{a\in A}cos(f(w),f(a)) -\\ 
& - mean_{b\in B}cos(f(w),f(b))
\end{aligned}
\end{equation} 

Alternatively, we can measure the size of the association via the Cohen's $d$:
\begin{equation}
    d = \frac{mean_{j\in J}s(j,A,B) - mean_{k\in K}s(k,A,B)}{std_{w\in J\cup K}s(w,A,B)}
\end{equation}

The authors found out that the representations they analysed contained several biased associations that resemble human cognitive biases (e.g., the association between flower/insects with pleasant/unpleasant respectively, male/female with career/family or white/black with tools/weapons). Nevertheless, it is not clear how many of those associations were present in the original training data and to what degree this is the responsibility of the CV models used for computing the representations. While this method for measuring biased associations is model agnostic, in the sense that it can be applied to any possible model, its results heavily depend on the learned representations and the employed models. 

\paragraph{Other}

\cite{Balakrishnan2020} developed a method for assessing \textit{algorithmic} biases in image classifiers via computing the causal relationship between the protected attribute and the output of a classifier. In doing so, they developed a method for intervening on the image attributes: they assumed to have a generator, such as a Generative Adversarial Network \citep{Goodfellow14}, that generates images from latent vectors. They also assumed to have learned hyper-planes in the latent space that separate different attributes. Sampling points along the direction orthogonal to an attribute's hyper-plane gives a modified version of the original image with respect to that attribute. The authors noted that such interventions are not entirely disentangled: for example, adding long hair to the images of white males also adds the beard, and changing the gender attributes to the images of black males adds earrings. This is probably due to datasets bias which is then detected as a side effect of the transformation described above. Note that, since such transformation is the result of geometric operations, it means that the bias is encoded in the geometry of the latent space. It would be interesting to study to what extent these manipulations of the latent space can be used as a bias exploration tool.

\subsection{Cross-dataset Bias Detection} 
\label{subsec:cross-dataset}
Methods in this category derive from the realisation that the issue of generalisation in CV might be due to bias. Researchers in the field of object detection are usually able to tell with fair accuracy which famous benchmark dataset an image comes from \citep{Torralba2011}. This means that each dataset carries some sort of ``signature" (bias) that makes the provenance of its images easily distinguishable and affects the ability of the models to generalise well. The methods described in the following aim to detect such signatures by comparing different datasets. Note that the works of \cite{Tommasi15, Lopez-Lopez2019} could also fit Section \ref{subsec:LatentRep}.

\paragraph{Cross-dataset generalisation}

\cite{Torralba2011} tested bias in object detection datasets by answering the following question: ``How well does a typical object detector trained on one dataset
generalize when tested on a representative set of other datasets, compared with its performance on the native test set?" The assumption here is that if the performance on the native test set is much higher it means the datasets exhibit some \emph{bias} that is learned by the object detector $f_{\mathcal{D}}$. Hence, let us consider a dataset of interest $\mathcal{D}$ and other $n$ benchmarks $\{\mathcal{B}_i\}$. They compare performance by computing:

\begin{equation}
\frac{1}{n}\sum_{i=1}^n
\frac{AP_{\mathcal{B}_i}(f_\mathcal{D})}{AP_{\mathcal{D}}(f_\mathcal{D})}
\end{equation}

The closer to 1 this score is, the more $f_\mathcal{D}$ generalises well. Note that if this score is low, we can say that the $f_\mathcal{D}$ does not generalise well and thus the dataset $\mathcal{D}$ is probably biased, while if the score is close to 1 we can say that the datasets share a similar representation of the visual world. Furthermore, the authors also proposed a test, which they presented as a toy experiment but which has been utilised in many other studies, called \textit{Name the dataset}. They trained a model to recognise the source dataset of a particular image: the greater the accuracy of this model, the more distinguishable and the more biased the datasets are. The methods described above have also been used by \cite{Tommasi15, Panda2018, Karkkainen2021}. 

\cite{Tommasi15} investigated the possibility of using CNN feature descriptors to address dataset bias. In particular, they replicated the experiments by \cite{Torralba2011} and \cite{Khosla12} (see next paragraph) using DeCAF features \citep{Donahue14}. Furthermore, they slightly changed the measure used by Torralba and Efros for the evaluation of cross-generalisation: 
\begin{equation}
    CD = \sigma(\frac{1}{100}(AP_\mathcal{D}(f_\mathcal{D}) - \frac{1}{n}\sum_{i=1}^{n} AP_{\mathcal{B}_i}(f_\mathcal{D})))
\end{equation}
where $\sigma()$ is the sigmoid function. 

\paragraph{Other}

\cite{Khosla12} proposed a method for both modelling and mitigating dataset bias. In particular, they train an SVM binary classifier on the union of a set of $n$ datasets $\mathcal{D}_i=\{(\mathbf{x}_j^i,y_j^i)\}_{j=1}^{s_i}$, where $|\mathcal{D}_i| = s_i$, $y_j^i \in \{-1,1\}$ is a common class among all the $n$ datasets, and where $\mathbf{x}_j^i$ are feature vectors extracted via some feature extraction algorithms. The problem is framed as a multi-task classification \citep{Evgeniou2004} where the algorithm learns $n$ distinct hyper-planes $\mathbf{w}_i\cdot\mathbf{x}$ where $\mathbf{w}_i = \mathbf{w} + \Delta_i$. The vector $\mathbf{w}$, which is common to each dataset, models the common representation of the visual world while the specific biases of each of the $\mathcal{D}_i$ datasets are modelled by the vectors $\Delta_i$. This is achieved via the following minimisation problem 
\begin{equation}\min_{\mathbf{w}, \Delta_i} \frac{1}{2}||\mathbf{w}||^2 + \sum_{i=1}^{n} \left[\frac{\lambda}{2}||\Delta_i||^2 + \mathcal{L}(\mathbf{w},\Delta_i) \right]
\end{equation}
where
\begin{equation}
\begin{aligned}
\mathcal{L}(\mathbf{w},\Delta_i)= & \sum_{j=1}^{s_i}(C_1 \min(1,y_j^i\mathbf{w}\cdot\mathbf{x_j^i}) +\\                 & + C_2\min(1,y_j^i(\mathbf{w}+\Delta_i)\cdot\mathbf{x_j^i})
\end{aligned}
\end{equation} 

and $\lambda$, $C_1$ and $C_2$ are hyper-parameters. The authors proved that studying the vectors $\Delta_i$ gives useful semantic information about each dataset bias (e.g., they discovered that Caltech101 \citep{FeiFei04} has a strong preference on the side view of cars).

\cite{Lopez-Lopez2019} also investigated the use of feature descriptors to discover biases. In particular, they wanted to understand how the images in different datasets are distributed after embedding them in the same lower-dimensional space. Given $n$ datasets, $\mathcal{D}_1$,...,$\mathcal{D}_n$ they sampled two sets of images for each of them $\mathcal{G}_1$,...,$\mathcal{G}_n$ and $\mathcal{P}_1$,...,$\mathcal{P}_n$, called respectively \textit{gallery sets} and \textit{probe sets}.
Then, they computed the latent space applying a pre-trained feature descriptor $f$ to the gallery sets. After that, they  computed the following probability:

\begin{equation}
    \mathbb{P}(\mathbf{x}_*\in \mathcal{D}_i|\mathbf{x}\in \mathcal{D}_i)
\end{equation}
where $\mathbf{x}$ is the feature vector of an image in the probe set $P_i$ and $\mathbf{x}_*$ is its nearest neighbour among the feature vectors of the images in the union of the gallery sets. If such probability is not $\frac{1}{n}$, it means that the nearest neighbours are not equally distributed among the $n$ datasets. Hence, there must be some \emph{selection} bias.

\subsection{Other Methods}
\label{subsec:other}

Other visual dataset bias discovery methods do not fit any of the three categories described above, and range from crowdsourcing frameworks to ad-hoc trained classification models.  

\paragraph{Model-based}
\cite{Model2015} proposed a simple method for addressing bias in object recognition datasets. They cropped a small central sub-image from each image in the original dataset. These cropped pictures were so small that humans could not recognise the objects in the pictures. Hence, if the attained performance of a model trained on such images was better than pure chance, the data contained distinguishable features spuriously correlated with the object categories. Then the data showed some kind of \emph{selection} or \emph{framing} (\emph{capture}) bias.

\cite{Thomas2019} studied the political framing bias of images by scraping images from online news outlets. Their idea was to train a semi-supervised tool for classifying images according to their political orientation. They labelled images according to the political orientation of the source. They also used information regarding the articles hosting the images by feeding the network with both the image and the document embedding of the article. Note that the proposed architecture incorporates textual information at training time but allows them to classify images without any additional information at testing time. Thus, this model can be used to understand if a specific image or a collection of images is biased towards a particular political faction. Moreover, the visual explanation of such a model could give semantic information on the political framing of the dataset. 

\cite{Clark2020} proposed to use an ensemble classification algorithm for mitigating bias. The idea is to train a low-capacity network (i.e., with a low number of parameters) together with a high-capacity one (i.e., with more than double the parameters) so that the former learns spurious correlations while the latter learns to classify the data in an unbiased way. While this difference in capacity and the ensemble training encourages the two models to learn different patterns, there is still the possibility for the high-capacity model to learn simpler patterns and hence bias. To avoid this, the two networks are trained to result in conditionally independent outputs. This will be an incentive for the ensemble to isolate simple and complex patterns. While the authors use this method with the only purpose of mitigating algorithmic bias, studying the lower-capacity model can give information on spurious correlations in the training dataset, highlighting selection/framing bias.

\cite{Lopez-Paz2017} proposed a method to discover ``causal signals" among objects appearing in image datasets. In particular, given a set of images $\mathcal{D} = \{d_i\}$, they score the presence of certain objects/attributes $A$, $B$ by using an object detector or a feature extractor, respectively. Then they apply a Neural Causation Coefficient (NCC, \citep{Lopez-Paz2015}) network, a neural network that takes a bag of samples $\{(a_i, b_i)_{i=1}^m\}$ as input and returns a score $s\in [0,1]$ where $s=0$ means that $A$ causes $B$ and $s=1$ means instead that $B$ causes $A$. Note that, while not every causal relationship is a source of bias, some might be. Moreover, such causal relationships can be thought of as a by-product of the selection of the subject. Hence this method can detect selection bias. 

\paragraph{Human-based}
\cite{Hu2020} proposed a non-automated approach for bias discovery. Their method consists of a three-step crowdsourcing workflow for bias detection (selection and framing, according to our categorisation). In order to avoid the complexity of free text description, in the first step, workers are presented with a batch of images and asked to describe the similarities among the images of the batch via a question-answer pair (e.g., if every image of the batch shows only white airplanes, and then there is some selection bias, the worker would label the batch with the following question-answer pair: What colour are the airplanes in the images? White). In a second step, each worker is asked to answer some of the questions collected in the first phase based on different batches to confirm the presence of such biases. Finally, the workers are asked to evaluate whether the statements are true in the real visual world or, on the contrary, constitute some biases (e.g., is it true that every airplane is white? If the answer is yes, this is not considered an instance of bias). Note that since this last step is based on ``common sense knowledge and subjective belief" \citep{Hu2020}, it heavily relies upon workers' backgrounds and biases. 

\subsection{Discussion}
\label{sec:prosCons}

The reduction to tabular data is a reasonable and effective way of discovering bias. These methods could leverage the great amount of work that has been already done in the field of Fair Machine Learning for tabular data. Nevertheless, it seems that for what concerns visual data, the methods used are relatively simplistic. Most of the works just look for balance in the protected attribute or compare the distribution with respect to other features. Furthermore, these methods heavily rely on labels that are either attached to the data or automatically extracted. Hence, any labelling bias affects such discovery methods and should be taken into account.

In a complementary way, the image representation methods reduce the problem to bias detection in a lower-dimensional space instead of reducing it to tabular data. While this could better capture the complexity of visual content, such methods are necessarily influenced by both the models used to compute the representation and the metric used to compute distances in it. Moreover, they are usually harder to apply since they need some kind of supervision for computing the space. 

Despite their historical importance, cross-dataset detection methods suffer from major issues. First, they are only applicable when several comparable datasets are available, which is not often practical. Second, they might help to unveil the presence of bias, but without further inspections, they cannot reveal what kind of bias it is. Note that, while these methods might be useful to get an idea of the existence of some biases in a visual dataset, they are of little or no use if we want to discover bias \textit{within} the dataset, for example, if we want to understand whether there are discriminatory differences between men and women in the same dataset.

Regarding those methods that do not fit any of the above-mentioned categories, we cannot outline common pros and cons as they are very problem/domain-specific.

\section{Bias-aware Visual Data Collection}
\label{sec:BiasFree}

\input{table_datasets}

In the following, we are going to describe some attempts to construct datasets where the existence of bias was taken into account during the dataset creation process. These datasets were constructed for specific purposes and were probably not thought of as universally bias-free datasets. Nonetheless, analysing which biases have been removed and which have been not might be helpful to understand the general challenge of bias in visual datasets. We summarise the content of this section in Table \ref{tab:datasets}. Furthermore, we propose a checklist for helping the collection of bias-aware visual datasets (Table \ref{tab:checklist}). Note that, while the previous section systematically reviewed the literature on bias discovery and quantification in visual data, this section is meant as a case study. Therefore, it is more limited in its scope.  

\subsection{Datasets} 
\label{subsec:bias-aware_datasets}

\cite{Buolamwini2018} released the Pilot Parliaments Benchmark (PPB) dataset. PPB is a \emph{face dataset} constructed by collecting the photos of members of six different national parliaments. The authors aimed to collect balanced data regarding the gender of the subjects and their skin colour. To do so, they selected three nations from African countries (Rwanda, Senegal, and South Africa) and three from European countries (Iceland, Finland, and Sweden) according to the gender parity rank\footnote{\url{https://data.ipu.org/women-ranking?month=6&year=2021}. Last visited 23.03.2022.} among their Members of Parliament (MP). The data have been labelled by three annotators (including the authors) according to (binary) gender appearance and the Fitzpatrick skin type (ranging from I to IV, these labels are used by dermatologist as a gold standard for skin classification). The definitive skin labels were provided by a board-certified dermatologist, while the definitive gender labels were also determined based on the title, prefix, or the name of the parliamentarians (note that using names as a proxy of gender can cause label bias \citep{Karimi16}). While the data collection process described above resulted in a much more balanced dataset compared to other famous benchmarks (Adience \citep{Eidinger14}; IJB-A \citep{Klare15}), it is still not free from possible biases. For example, the selection process targets a small number of African and northern European countries to ensure gender and skin tone balance. Nevertheless, it completely excludes, for instance, Asian and South American countries. Moreover, MPs are likely to be middle-aged people, which could also exclude young and older people from the selection. On the framing bias side, different countries might have different standards/dress codes for their MPs' official portraits, which could turn into some biases as well. 

\cite{Karkkainen2021} collected a \emph{face dataset} emphasising, in particular, the balance in terms of age, gender, and race. They relied on a crowdsourcing workflow for annotating the images. In particular, they asked three different workers to label the images according to gender, age group, and race. They kept the label if there was a 2/3 accordance. Otherwise, they would have proposed the image to other three workers and discarded it if again it resulted in three different judgements. We can spot two sources of label and selection bias, respectively: first, we cannot be sure that the workers are able to determine the three labels homogeneously across every sub-group \footnote{see the interesting Twitter thread from Sacha Costanza-Chock (@schock), \url{https://twitter.com/schock/status/1346478831255789569} (Last visited 23.03.2022) for an on-point discussion about this issue}; second, discarding the photos the workers cannot agree on might result in the missed selection of a certain group of people whose characteristics are difficult to determine for the workers. Finally, the taxonomy of races used by the authors (White, Black, Indian, East Asian, Southeast Asian, Middle East, and Latino) already introduces a form of label bias. While it is derived from the taxonomy commonly used by the US Census Bureau and might be descriptive of the composition of the US population, it hardly captures the complexity of human diversity.

Diversity in Face (DiF) \citep{Merler2019} and KANFaces \citep{Georgopoulos2020} are two \emph{face datasets} that try to address the issue of bias by ensuring as much diversity as possible using the diversity measures proposed by \cite{Merler2019} that we reviewed in Section \ref{sec:Discovery}. The attributes they control the diversity for are: age, gender, skin tone, a set of craniofacial ratios, and pose. The authors also took into account a metric of illumination. By collecting diverse data, the authors try to avoid selection bias (in the case of age, gender, and skin tone) and some framing bias (in the case of pose and illumination).

\cite{Barbu2019} made an attempt to avoid framing bias in large-scale \emph{object detection datasets}. In particular, they added controls for object rotations, viewpoints and background by asking crowd workers to photograph objects in their homes in a natural setting according to the instructions given by the authors. While this resulted in a much more diverse dataset (the authors used ImageNet \citep{Russakovsky15} as a comparison), because of the above controls, the objects appear only in indoor context, are rarely occluded, and are often centre-aligned. Thus, it seems that specific framing biases have been avoided, while the collection procedure has introduced others. Also, the authors removed some classes from the dataset for reasons that range from privacy concerns (e.g., ``people'') or because they were not easy to move and photograph in different settings (e.g., ``beds''). In principle, this might generate some selection bias (more specifically, negative class bias) since the absence of those objects could make the negative classes less representative.    

\cite{Wu2020} collected two benchmark datasets: the Inclusive Benchmark Database (IBD), and Non-binary Gender Benchmark Database (NGBD)\footnote{While collecting inclusive data is a noble purpose, the use the authors made of their dataset (extending gender classification algorithms to non-binary genders) should be strongly discouraged as it might be used to target already marginalised groups of people.}. IBD contains 12,000 images of 168 different subjects, 21 of whom identify as LGBTQ. The geographic provenance of the subjects in the dataset is balanced. NGBD contains 2,000 images of 67 unique subjects. The subjects are public figures whose gender identity is known. Thus, the database contains multiple gender identities (namely: \textit{non-binary}, \textit{genderfluid}, \textit{genderqueer}, \textit{gender non-conforming}, \textit{agender}, \textit{gender neutral}, \textit{gender-less}, \textit{third gender}, and \textit{queer}). The authors themselves identify two major risks of label bias: first, ``Gender identity has its multifaceted aspects that a simple label could not categorize" \citep{Wu2020} (the authors identify the problem of modelling gender as a continuum as a direction for future work); and, second, ``Gender is a complex socio-cultural construct and an internal identity that is not necessarily tied to physical appearances." \citep{Wu2020}. 

\cite{Hazirbas2021} proposed the Casual Conversations Dataset for evaluating the performance of CV models across different demographic categories. Their dataset is composed of 3,011 subjects and contains over 45,000 videos, with an average of 15 videos per person. The videos were recorded in multiple US states with a diverse set of adults in various age, gender and apparent skin tone groups. This work represents probably the greatest effort to build a balanced dataset addressing both selection and framing bias (in the form of the illumination of videos). Nevertheless, some forms of imbalance remain. For example, most videos present bright lighting conditions; and most subjects are labelled as either male or female (with just the 0.1\% of the participants that identify as ``Others" and 2.1\% whose gender is unknown). The label bias that the use of categories such as gender, age, and race could create is overcome by asking the participants their age and gender and using the Fitzpatrick Skin Type instead of race. Nonetheless, the authors declare that there are ``videos in which two subjects are present simultaneously" but that they provide only one set of labels which might also be a form of label bias. Last, we note that the subjects are all from the US, which represents a serious selection bias as the US population hardly represents the whole of humankind. 

\subsection{Discussion}
\label{subsec:discussion_checklist}

As highlighted by the cases analysed in the previous section, dealing with bias in visual data is not a trivial task. In particular, collecting bias-aware visual datasets might be incredibly challenging. Thus, we propose a checklist (Table \ref{tab:checklist}) to help scientists and practitioners spot and make explicit possible biases in the data they collect or use. Furthermore, we add a brief discussion on pre-processing mitigation strategies. Note that bias mitigation is an area on its own and would require an entire review to be adequately discussed. Therefore, we will just provide the interested reader with some hints and references.

Our checklist is inspired by previous work on documentation and reflexive data practices \citep{Gebru2018, Miceli21}, but adds several questions specific to selection, framing, and label bias because they have their own peculiarities and must be analysed separately. We start with a general set of questions on the dataset purpose and the collection procedure. Then, we continue with questions specific to selection, framing, and label bias. In particular, we ask whether the selection of subjects generates any imbalance or lack of diversity, whether there are spurious correlations or harmful framing, whether fuzzy categories are used for labelling, and whether the labelling process contributes to inserting biases. 

\input{table_checklist}

We want to stress that careful data collection and curation are probably the most effective mitigation strategies (see, for example, the mitigation insights described in \citep{Wang2020}) for all the three types of bias presented in \ref{sec:CVBias}. However, selection bias seems to be the easiest to mitigate using standard pre-processing techniques such as re-sampling or re-weighting. Note that several image datasets have a long-tail distribution of objects, and hence pre-processing mitigation techniques have to take that into account (see \citep{Zhang21}), especially when many objects can co-occur in the same image. When the label bias is generated by poor operationalisation of unobservable theoretical construct \citep{Jacobs21}, the entire labelling must be reconsidered. Otherwise, when bias is caused by the presence of synonyms or incorrect labelling, an effective data cleaning could help the mitigation. When it appears in the form of capture bias, framing bias can be mitigated by either re-sampling techniques that ensure diversity (in pose and lighting, for example) or via data augmentation. When the framing bias relates to the messages carried by the images, automatic pre-processing strategies are probably less effective as the way an image is interpreted and the message it conveys are not universal, which gives rise to the need for careful human inspection and analysis.

\section{Conclusions and Research Outlook}
\label{sec:Conclusions}

The aim of this survey was threefold: first, to provide a description of different types of bias and to illustrate the processes through which they affect CV applications and visual datasets; second, to perform a systematic review of methods for bias discovery in visual datasets; and, third, to describe existing attempts to collect visual datasets in a bias-aware manner. We showed how the problem of bias is pervasive in CV; It accompanies the whole life cycle of visual content, involves several actors, and re-enters the life cycle through biased CV algorithms. One of our major contributions has been to provide a detailed description of the different manifestations (selection, framing, and label) of bias in visual data, along with several examples. We also went further by providing a sub-categorisation (Table \ref{tab:sub_biases}) which also includes several categories of bias that are commonly described in Statistics, Health studies, or Psychology adapting them to the context of visual data. 

The systematic review in Section \ref{sec:Discovery} allowed us to draw some consideration on the state of the art in bias discovery methods for visual data and to outline some possible future streams of research. The vast majority of them used the strategy of reducing the problem to tabular data (Section \ref{subsec:Reduction}). While it seems a natural option as it allows leveraging the wealth of techniques from the Fair Accountable and Transparent Machine Learning (FATML) literature, using different representations of visual data would certainly open new unexplored research directions. For example, representing an image as a scene graph \citep{Johnson2015}, would allow the use of bias detection techniques used for rankings and graphical data \citep{krasanakis2020graphfairness,Pitoura21}. Moreover, since scene graphs can be thought of as small-scale knowledge graphs, this representation would allow both the integration of additional knowledge (e.g., from Wikipedia) and the use of methods for bias detection in knowledge graph embeddings \citep{Bourli20}.

\cite{Balakrishnan2020} showed that a latent representation of data encapsulates bias in its geometry. This was also noted by \cite{Bolukbasi2016} in their work on bias in word embeddings. Nonetheless, the study of this relationship between the geometry of latent spaces and bias is usually limited to variations of cosine similarity. Hence, a promising research direction would be to exploit more sophisticated tools for studying such relationships, e.g., Topological Data Analysis \citep{Chazal2021}. 

Some of the most influential works on bias detection in images, e.g., \citep{Torralba2011}, focused on object detection datasets. In such datasets, the long-tail distribution of objects is very common and evaluating bias in these datasets is difficult because the long tail distracts from the focal points. Moreover, the co-occurrence of different objects has obvious implications for the effectiveness of mitigation methods \citep{Zhang21} for imbalanced data (e.g., oversampling, under-sampling), and therefore it should be studied more deeply. 

Most of the reviewed papers focus on images. Only few of them study bias in videos. This opens the road to many possibilities (for example, applying methods from time-series and multimodal analysis to the bias detection domain). Furthermore, we noted that selection and framing bias are the two types of bias that are more commonly taken into account. Nevertheless, label bias is often very concerning (Section \ref{sec:labelbias}) and hence more research is needed to fill this gap. 

Similarly, for obvious simplicity reasons, most related works concentrate on a single protected attribute (an exception to this is the work of \cite{Buolamwini2018}). Hence, multi-attribute fairness studies would enrich the literature in this respect. We also noted that most works we reviewed in Section \ref{sec:BiasFree} focus on facial data. This is probably due to the concerning applications of face detection/recognition algorithms. Therefore, there is room for improving datasets and establishing data collection practices in other fields, including medical imaging, self-driving cars, etc. 

We finally reviewed several attempts to collect bias-aware data in Section \ref{sec:BiasFree} and concluded that there is no such thing as bias-free data. Hence, it is of utmost importance, along with the development of reliable bias discovery tools, that researchers and practitioners become aware of the biases of the datasets they collect and make them explicit in a standardised way (see, for example, \citep{Gebru2018, Miceli21}). In Table \ref{tab:checklist}, we outline a checklist for the collection of visual data. We believe that having such a guide will help practitioners and scientists spot possible causes of bias, collect data that is as less biased as possible, and be aware of such biases during their analysis. 

\paragraph{Acknowledgements} We would like to thank Alaa Elobaid, Miriam Fahimi and Giorgos Kordopatis-Zilos for their feedback and fruitful discussions. This work has received funding from the European Union’s Horizon 2020 research and innovation programme under Marie Sklodowska-Curie Actions (grant agreement number 860630) for the project ``NoBIAS - Artificial Intelligence without Bias" and under grant agreement number 951911 for the project ``AI4Media - A European Excellence Centre for Media, Society and Democracy''. This work reflects only the authors’ views and the European Research Executive Agency (REA) is not responsible for any use that may be made of the information it contains.

\bibliography{Bib}{}
\bibliographystyle{plainnat} 

\end{document}

%% file: table_sub_biases.tex
\begin{table*}
    \centering
    \small
    \begin{adjustbox}{angle=0}
  
    \begin{tabular}{|l|p{75mm}|c|c|c|}
    \cline{3-5}
    \multicolumn{2}{c|}{}&\multirow{4}{*}{\rotatebox{270}{\textbf{Selection}}}&\multirow{4}{*}{\rotatebox{270}{\textbf{Framing}}}&\multirow{4}{*}{\rotatebox{270}{\textbf{Label}}}\\
    
    \multicolumn{2}{c|}{}&&&\\
    \multicolumn{2}{c|}{}&&&\\
    \multicolumn{2}{c|}{}&&&\\
    \cline{1-2} 
    \textbf{Name} & \textbf{Description} &  &  &   \\
    \hline
    Sampling bias$^*$ & Bias that arises from the sampling of the visual data. It includes class imbalance. & $\bullet$ &  &   \\
    
    Negative set bias \citep{Torralba2011} & When a negative class (say \texttt{non-white} in a \texttt{white/non-white} categorisation) is not representative enough. & $\bullet$  &  & $\bullet$  \\
    
    Availability bias$^\dagger$ & Distortion arising from the use of the most readily available data (e.g., using search engines).& $\bullet$ &  & \\ 
    
    Platform bias & Bias that arises as a result of a data collection being carried out on a specific digital platform (e.g., Twitter, Instagram, etc.). & $\bullet$ &  &  \\
    
    Volunteer bias$^\dagger$ & When data is collected in a controlled setting instead of being collected in-the-wild, volunteers that participate in the data collection procedure may differ from the general population. & $\bullet$ &  &   \\
    
    Crawling bias & Bias that arises as a result of the crawling algorithm/system used to collect images from the Web or with the use of an API (e.g., the keywords used to query an API, the seed websites used in a crawler). & $\bullet$ & $\bullet$ &  \\
    
    Spurious correlation & Presence of spurious correlations in the dataset that falsely associate a certain group of subjects with any other features. & $\bullet$ & $\bullet$ &   \\
    
    Exclusion bias$^*$ & Bias that arise when the data collection excludes partly or completely a certain group of people. & $\bullet$ & $\bullet$ &   \\
    
    Chronological bias$^\dagger$ & Distortion due to temporal changes in the visual world the data is supposed to represent. & $\bullet$  & $\bullet$ & $\bullet$ \\
    
    Geographical bias \citep{Shankar2017}& Bias due to the geographic provenance of the visual content or of the photographer/video maker (e.g., brides and grooms depicted only in western clothes). & $\bullet$  & $\bullet$ &   \\
    
    Capture bias \citep{Torralba2011} & Bias that arise from the way a picture or video is captured (e.g., objects always in the centre ,exposure, etc.). &  & $\bullet$  &  \\
    
    Apprehension bias$^\dagger$ & Different behaviour of the subjects when they are aware of being photographed/filmed (e.g., smiling).  &  & $\bullet$  & \\
    
    Contextual bias \citep{Singh2020} & Association between a group of subjects and a specific visual context (e.g., women and men respectively in household and working contexts) &  & $\bullet$ &   \\ 
    
    Stereotyping$^\mathsection$ & When a group is depicted according to stereotypes (e.g., female nurses vs. male surgeons).  &  & $\bullet$  &   \\
    
    Measurement bias \citep{Jacobs21} & Every distortion generated by the operationalisation of an unobservable theoretical construct (e.g., race operationalised as a measure of skin colour). &  &  & $\bullet$   \\ 
    
    Observer bias$^\dagger$ & Bias due to the way a annotator records the information. &  &  & $\bullet$   \\
    
    Perception bias$^\dagger$ & When data is labelled according to the possibly flawed perception of a annotator (e.g., perceived gender or race) or when the annotation protocol is not specific enough or is misinterpreted. &  &  &  $\bullet$  \\ 
    
    Automation bias$^\mathsection$ & Bias that arises when the labelling/data selection process relies excessively on (biased) automated systems. & $\bullet$ &  &  $\bullet$  \\
    
    \hline
    \end{tabular}
    \end{adjustbox}
    \caption{A finer-grained categorisation of biases described in Section \ref{sec:CVBias}. The bullets represent which types of bias they are a subcategory of (Note that there are overlaps as different kind of bias can co-occur). The definitions are adapted from: ($\dagger$) \url{https://catalogofbias.org/biases/}; (*) \url{https://en.wikipedia.org/wiki/Sampling_bias}; ($\mathsection$) \url{https://en.wikipedia.org/wiki/List_of_cognitive_biases}; and \citep{Torralba2011, Shankar2017, Singh2020, Jacobs21}.
    }
    \label{tab:sub_biases}
\end{table*}

%% file: table_notation.tex
\begin{table*}
    \centering
    \small
    \begin{adjustbox}{angle=0}
    \begin{tabular}{|l|l|}
    \hline
    \textbf{Symbol} & \textbf{Description}\\
    \hline
    $\mathcal{D}$ & Cursive capital letters denotes visual datasets\\

    $f_\mathcal{D}(\cdot)$ & A model $f$ trained on the dataset $\mathcal{D}$ \\  
    
     $\mathbf{w}$ & Bold letters denotes vectors\\
    
    $AP$ & Average Precision-Recall (AP) score\\
    
    $AP_\mathcal{B}(f_\mathcal{D})$ & AP score of the model $f_{\mathcal{D}}$ when tested on the dataset $\mathcal{B}$\\
    
    $||\cdot ||_2$ & L2-norm\\
    
    $|\mathcal{D} |$ & The number of elements in the dataset $\mathcal{D}$\\
    
    $\sigma(\cdot)$ & Sigmoid function \\
    
    $\mathbf{1}[\cdot]$  & Indicator function\\
    
    $\mathbb{P}(\cdot)$ & Probability\\
    
    $\mathbb{P}(\cdot|\cdot)$ & Conditional probability\\
    
    $H(\cdot)$ & Shannon entropy\\
    
    $H(\cdot|\cdot)$ & Conditional Shannon entropy\\
    
    $I(\cdot, \cdot)$ & Mutual information\\
    
    $D(\cdot)$ & Simpson $D$ score\\
    
    $ln(\cdot)$ & Natural logarithm\\
    
    $mean(\cdot)$ & Arithmetic mean\\
    
    \hline
    \end{tabular}
    \end{adjustbox}
    \caption{Brief summary of the notation used in Section \ref{sec:Discovery}.}
    \label{tab:notation}
\end{table*}

%% file: table_metrics.tex
\begin{table*}
    \centering
    \small
    \begin{adjustbox}{angle=90}
    \begin{tabular}{|l|l|l|l|c|c|c|l|}
    \cline{5-7}
    \multicolumn{4}{c|}{}&\multirow{4}{*}{\rotatebox{270}{\textbf{Selection}}}&\multirow{4}{*}{\rotatebox{270}{\textbf{Framing}}}&\multirow{4}{*}{\rotatebox{270}{\textbf{Label}}}&\multicolumn{1}{}{}\\
    
    \multicolumn{3}{}{}&&&&\\
    \multicolumn{3}{}{}&&&&\\
    \multicolumn{3}{}{}&&&&\\
    \cline{2-4} \cline{8-8}
    \multicolumn{1}{c|}{}& \textbf{No.} & \textbf{Paper} & \textbf{Year} &  &  &  & \textbf{Type of measures/methods}  \\
    \hline

    \multirow{13}{*}{\textbf{Reduction to tabular data}}& 1& \cite{Dulhanty2019} & 2019 & $\bullet$ &  $\bullet$ & & Count; Demographic parity\\
    
    & 2& \cite{Yang2020} & 2020 &$\bullet$ & $\bullet$ & $\bullet$ & Count; Demographic parity\\

    & 3& \cite{Zhao2017} & 2017 & $\bullet$& $\bullet$& & Demographic parity\\
    
    & 4& \cite{Shankar2017} & 2017 & $\bullet$& & & Count\\
    
    & 5& \cite{Buolamwini2018} & 2018 &$\bullet$ & & & Count\\
    
    & 6& \cite{Merler2019} & 2019 &$\bullet$ & & & Entropy-based; Information theoretical\\
    
    & 7& \cite{Panda2018} & 2018 &$\bullet$ & $\bullet$ & & Entropy-based\\
    
    & 8& \cite{Kim2019} & 2019 & $\bullet$& $\bullet$ & & Information theoretical\\
    
    & 9& \cite{Wang2019} & 2019 & $\bullet$ & $\bullet$ & & Dataset leakage\\
    
    & 10& \cite{Wachinger2021} & 2021 & $\bullet$ & & & Causality\\
    
    & 11& \cite{Jang2019} & 2019 & &$\bullet$ & & 4 different measures\\
    
    & 12& \cite{Wang2020} & 2020 &$\bullet$ &$\bullet$ &$\bullet$ & 13 different measures\\
    \hline
    
    \multirow{4}{*}{\textbf{Biased image representation}}& 13& \cite{Karkkainen2021} & 2021 &$\bullet$ & & & Distance-based\\
    
    & 14& \cite{Steed21} & 2021 & &$\bullet$ & & Distance-based\\
    
    & 15& \cite{Balakrishnan2020} & 2020 & &$\bullet$ & & Interventions\\
    \hline
    
    \multirow{5}{*}{\textbf{Cross-dataset bias detection}}& 16& \cite{Torralba2011} & 2011 & $\bullet$ & $\bullet$ & & Cross-dataset generalisation\\
    
    & 17& \cite{Tommasi15} & 2015 & $\bullet$ & $\bullet$ & & Cross-dataset generalisation\\
    
    & 18& \cite{Khosla12} & 2012 & $\bullet$ & $\bullet$& & Modelling bias\\
    
    & 19& \cite{Lopez-Lopez2019}& 2019 & $\bullet$ & & & Nearest neighbour in a latent space\\
    \hline
    
    \multirow{4}{*}{\textbf{Other}}& 20& \cite{Model2015} & 2015 & $\bullet$ & $\bullet$ & & Model-based\\
    
    & 21& \cite{Thomas2019} & 2019 & & $\bullet$ & & Model-based\\
    
    & 22& \cite{Clark2020} & 2020 & $\bullet$ & $\bullet$ & & Modelling bias\\
    
    & 23& \cite{Lopez-Paz2017} & 2017 &$\bullet$ & & & Causality\\
    
    & 24& \cite{Hu2020} & 2020 &$\bullet$ & $\bullet$& & Crowd-sourcing\\

    \hline
        
    \end{tabular}
    \end{adjustbox}
    \caption{Summary of the collected material.}
    \label{tab:summary}
\end{table*}

%% file: table_datasets.tex
\begin{table*}
    \centering
    \small
    \begin{adjustbox}{angle=90}
    \begin{tabular}{|p{35mm}|l|l|p{28mm}|l|p{35mm}|c|c|c|}
    
        \cline{7-9}
        \multicolumn{6}{c|}{}&\multirow{4}{*}{\rotatebox{270}{\textbf{Selection}}}&\multirow{4}{*}{\rotatebox{270}{\textbf{Framing}}}&\multirow{4}{*}{\rotatebox{270}{\textbf{Label}}}\\
    
        \multicolumn{6}{c|}{}& & &\\
        \multicolumn{6}{c|}{}& & &\\
        \multicolumn{6}{c|}{}& & &\\
        \cline{1-6} 
        \textbf{Paper} & \textbf{Year} & \textbf{Size} & \textbf{Protected attributes} & \multicolumn1{p{15mm}|}{\textbf{Pre-existing Content}} &  \multicolumn1{p{30mm}|}{\textbf{Labelling Process}}& & &\\
        \hline
        
        \cite{Buolamwini2018} & 2018 & 1.2K images & Binary gender; skin tone & Yes & Human annotators; experts; additional information & $\bullet$ & & $\bullet$\\
         
        \cite{Karkkainen2021} & 2021 & 108.5K images &  Age; binary gender; race & Yes & Crowd workers & $\bullet$ & &\\
         
        \cite{Merler2019} & 2019 & 970K images & Age; Binary gender; skin tone & Yes &  Machine annotators; crowd workers& $\bullet$ & $\bullet$ &\\
         
        \cite{Georgopoulos2020} & 2020 &\multicolumn1{p{18mm}|}{41K images; 44K videos} & Age; binary gender & Yes & Human annotators; machine annotators & $\bullet$ & $\bullet$ &\\
         
         \cite{Barbu2019} & 2019 & 50K images & - & No & - & &$\bullet$ &\\
         
        \cite{Wu2020} (Inclusive Benchmark)& 2020 & 12K images & Binary gender; race & Yes & Additional information & $\bullet$ & &\\
         
        \cite{Wu2020} (Non-binary Gender Benchmark) & 2020 & 2K images & Non-binary gender; race & Yes & Additional information & $\bullet$ & &$\bullet$\\
         
         \cite{Hazirbas2021} & 2021 & 45.1K videos & Age; Non-binary gender; skin tone & No & Human annotators; self-provided labels & $\bullet$ & $\bullet$ & $\bullet$\\
         
         \hline
    
    \end{tabular}
    \end{adjustbox}
    \caption{Summary of the datasets described in Section \ref{sec:BiasFree}.}
    \label{tab:datasets}
\end{table*}

%% file: table_checklist.tex
\begin{table*}[h!]
    \centering
    \small
    \begin{adjustbox}{angle=0}
    \begin{tabular}{|l|p{100mm}|}
    \cline{2-2}
    \multicolumn{1}{c|}{}& \textbf{Description}\\
    \hline
    
    \multirow{3}{*}{\textbf{General}}& What are the purposes the data is collected for? \citep{Gebru2018} \\
    
    & Are there uses of the data that should be discouraged because of possible biases? \citep{Gebru2018}\\

    & What kind of bias can be inserted by the way the collection process is designed? \citep{Gebru2018}\\
    
    \hline
    
    
    \multirow{7}{*}{\textbf{Selection bias}}& Do we need balanced data or statistically representative data?\\
    
    & Are the negative sets representative enough?\\
    
    & Is the dataset representative enough?\\
    
    & Is there any group of subjects that is systematically excluded from the data? \\
    
    & Do the data come from or depict a specific geographical area? \\
    
    & Does the selection of the subjects create any spurious associations?\\
    
    & Will the data remain representative for a long time?\\
    \hline
    
    \multirow{6}{*}{\textbf{Framing bias}}& Are there any spurious correlation that can contribute to framing different subjects in different ways?\\
    
    & Is there any biases due to the way images/videos are captured?\\
    
    & Did the capture induce some behaviour in the subjects (e.g. smiling when photographed)?\\

    & Are there any images that can possibly convey different messages depending on the viewer?\\
    
    & Are subjects of a certain group depicted in a particular context more often than others?\\
    
    & Do the data agree with harmful stereotypes?\\
    \hline
    
    \multirow{5}{*}{\textbf{Label bias}}& If the labelling process relies on machines: have their biases been taken into account?\\
    
    & If the labelling process relies on human annotators: is there an adequate and diverse pool of annotators? Have their possible biases been taken into account?\\

    & If the labelling process relies on crowd sourcing: are there any biases due to the workers' access to crowd sourcing platforms? \\
    
    & Do we use fuzzy labels? (e.g., race or gender)\\
    
    & Do we operationalise any unobservable theoretical constructs/use proxy variables? \citep{Jacobs21}\\
    \hline
        
    \end{tabular}
    \end{adjustbox}
    \caption{Checklist for bias-aware visual data collection}
    \label{tab:checklist}
\end{table*}